\documentclass[runningheads]{llncs}
\usepackage{graphicx}
\usepackage{multirow}
\usepackage{algorithm}
\usepackage{amsmath}
\usepackage{amssymb}
\usepackage{booktabs}
\usepackage{hyperref}
\usepackage[misc]{ifsym}
\usepackage{hyperref}

\begin{document}
%
\title{M-FLAG: Medical Vision-Language Pre-training with Frozen Language Models and Latent Space Geometry Optimization}
\titlerunning{Medical Vision-Language Pre-training}
%
%
\author{$\text{Che Liu}^{1,2}\textsuperscript{(\Letter)}$, $\text{Sibo Cheng}^{2,3}$, $\text{Chen Chen}^{3,5}$,\\
$\text{Mengyun Qiao}^{2,4}$, $\text{Weitong Zhang}^{3}$, $\text{Anand Shah}^{6,7}$,\\
$\text{Wenjia Bai}^{2,3,4}$, $\text{Rossella Arcucci}^{1,2}$}
\authorrunning{C. Liu et al}
\institute{$^{1}$Department of Earth Science and Engineering, Imperial College London, UK\\
            $^{2}$Data Science Institute, Imperial College London, UK\\
            $^{3}$Department of Computing, Imperial College London, UK\\
            $^{4}$Department of Brain Sciences, Imperial College London, UK\\
            $^{5}$Department of Engineering Science, University of Oxford, UK\\
            $^{6}$Department of Infectious Disease Epidemiology, Imperial College London, UK\\
            $^{7}$Royal Brompton and Harefield Hospitals, UK\\
            che.liu21@imperial.ac.uk}

\maketitle              
\begin{abstract}
Medical vision-language models enable co-learning and integrating features from medical imaging and clinical text. However, these models are not easy to train and the latent representation space can be complex. Here we propose a novel way for pre-training and regularising medical vision-language models. The proposed method, named \textbf{M}edical vision-language pre-training with \textbf{F}rozen language models and \textbf{L}atent sp\textbf{A}ce \textbf{G}eometry optimization (M-FLAG), leverages a frozen language model for training stability and efficiency and introduces a novel orthogonality loss to harmonize the latent space geometry. We demonstrate the potential of the pre-trained model on three downstream tasks: medical image classification, segmentation, and object detection. Extensive experiments across five public datasets demonstrate that M-FLAG significantly outperforms existing medical vision-language pre-training approaches and reduces the number of parameters by 78\%. Notably, M-FLAG achieves outstanding performance on the segmentation task while using only 1\% of the RSNA dataset, even outperforming ImageNet pre-trained models that have been fine-tuned using 100\% of the data. The code can be found in \href{https://github.com/cheliu-computation/M-FLAG-MICCAI2023}{https://github.com/cheliu-computation/M-FLAG-MICCAI2023}.

\end{abstract}
\section{Introduction}
Deep learning has made significant progress in medical computer vision~\cite{chai2021deep,esteva2021deep} but requires large annotated datasets, which are often difficult to obtain. Self-supervised learning (SSL) offers a solution by utilizing large unannotated medical image sets. It also enables vision-language pre-training (VLP), which learns representations for both imaging and text data and their relationships~\cite{clip,wan2023med,chen2023generative}. Several recent medical VLP approaches such as ConVIRT~\cite{convirt}, GLoRIA~\cite{huang2021gloria}, and MGCA~\cite{mgca} have shown the effectiveness of model pre-training with medical images and radiology reports together, which outperformed the conventionally pre-trained models using image only in downstream tasks~\cite{convirt}.
However, training such models is not an easy task as they require extensive resources for training both vision and language models. In particular, most VLP approaches are based on pre-trained BERT~\cite{devlin2018bert,li2023frozen}, whose parameters are 5 times larger than a standard ResNet50~\cite{resnet}. This indicates high computational cost, as well as training complexity and potential instability in joint training~\cite{izsak2021train}. On the other hand, previous works~\cite{clip,convirt,mgca} suggest a training strategy that forces image latent space to match language latent space, which can be sub-optimal with latent space collapse problem~\cite{jing2021understanding}, reducing its performance for downstream tasks~\cite{zhu2017toward}. In this work, we would like to answer the following two questions: \textit{(1) Is it necessary to tune pre-trained language models for medical VLP? (2) How to regularize the latent space in pre-training?}

We propose a novel VLP framework named \textbf{M}edical vision-language
pre-training with \textbf{F}rozen language models and \textbf{L}atent sp\textbf{A}ce Geometry
optimization method (M-FLAG). Different from most existing VLP approaches, M-FLAG is computationally efficient as it only requires training the vision model, while keeping the language model frozen. To harmonize the latent spaces in vision and language models, we relax the visual-language alignment objective with a orthogonality loss to alleviate the latent space collapse problem. The main contributions of this work include:
\textbf{(1)} To the best of our knowledge, this is the first work to explore the collapsed latent space problem in medical VLP.
\textbf{(2)} A novel and effective VLP framework is proposed to alleviate the collapsed latent space problem by explicitly optimizing the latent geometry towards orthogonal using our orthogonality loss in addition to the visual-language alignment loss, encouraging the in-dependency between latent variables and maximizing its informativeness for downstream tasks.
\textbf{(3)} M-FLAG consistently outperforms existing medical VLP methods on three downstream tasks: medical image classification, segmentation, and object detection, while reducing 78\% trainable parameters due to the frozen language model strategy.

\noindent\textbf{Related works:} To connect vision and language modalities, the idea of VLP was proposed in CLIP~\cite{clip}, which involves learning mutual knowledge from two modalities by maximizing their feature similarity. CLIP~\cite{clip} and later FLIP~\cite{flip} focus on learning cross-representation in natural language and images. However, there is a significant lack of research in the medical domain due to the complexity of the medical text and the limited availability of large-scale paired medical image-text datasets. Recently, ConVIRT~\cite{convirt}, GLoRIA~\cite{huang2021gloria}, and MGCA~\cite{mgca} have made notable progress in aligning medical text and images. These methods require significant computational resources and are sometimes limited by the collapse issue of the latent space.

It has been suggested that optimal vision and language latent spaces should be of different geometry~\cite{fu2022latent} and latent space uniformity is considered an essential indicator to evaluate the success of learning~\cite{wang2020understanding}. Yet, most existing VLP approaches rigorously align the vision latent space to the language space without considering the latent space geometry, which may lead to latent space collapse.
As pointed out by~\cite{jing2021understanding}, latent space collapse indicates significant information loss, which can crucially affect the robustness of the pre-trained model on downstream tasks when transferring the model to unseen domains~\cite{chen2022perfectly}. To solve this problem, contrastive learning-based approaches can be used to spread visual features over the unit sphere with good uniformity~\cite{moco,simclr}. However, it requires a large number of negative samples in each training batch, which inevitably increases computational costs. Differently, here we address this problem by employing a orthogonality loss, which directly aligns the geometry of the latent space towards a uniform hypersphere to tackle the collapse problem. 

\section{Methods}
\begin{figure}[t]
\centering
\includegraphics[width=1.0\textwidth]{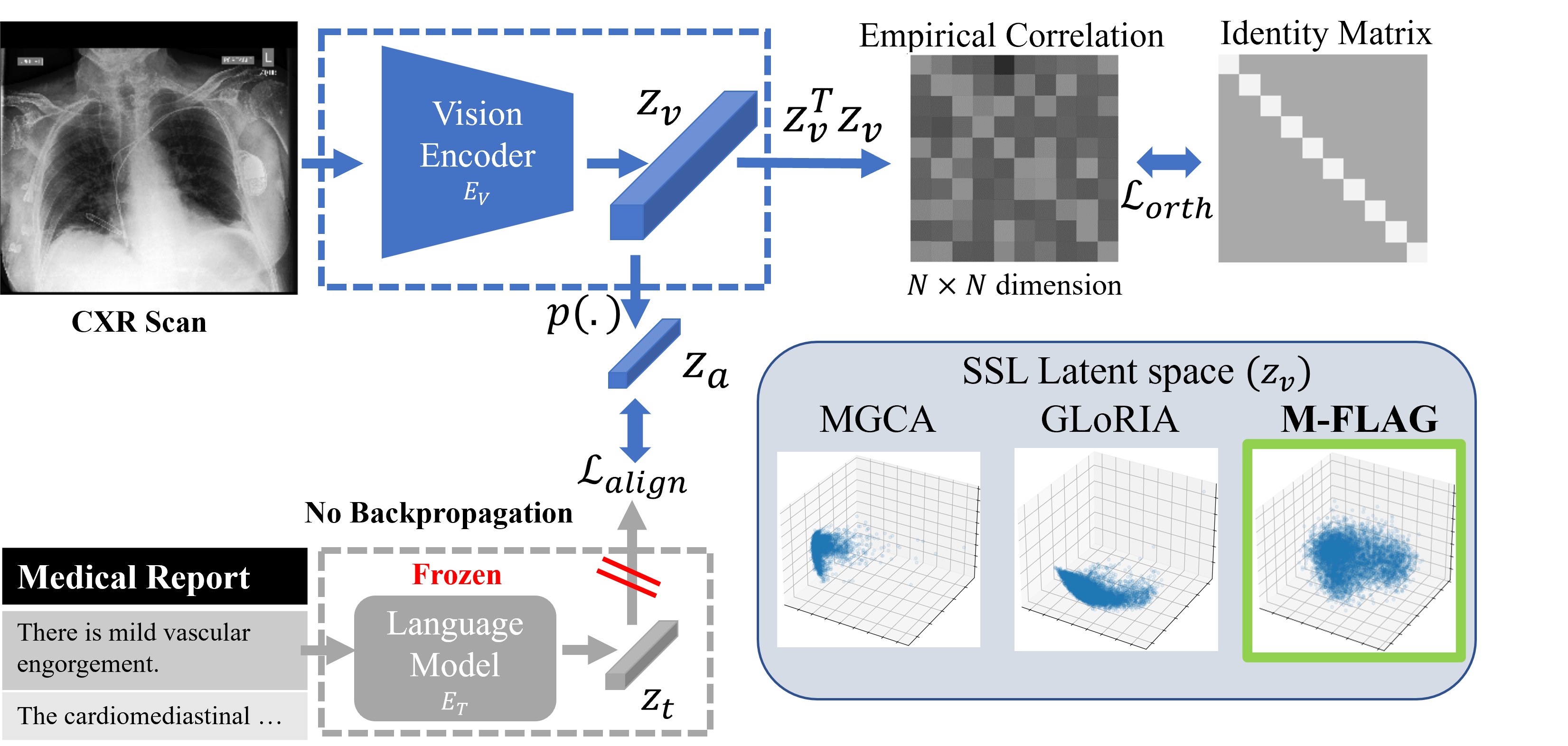}
\caption{\textbf{M-FLAG overview.} M-FLAG consists of a vision encoder $E_V$ for learning vision latent $z_v$, a \textit{frozen} language model $E_T$ for extracting medical text latent $z_t$, and a projector $p(\cdot)$ to map $z_v$ to $z_a$ for alignment with $z_t$. M-FLAG employs an alignment loss $\mathcal{L}_{align}$ for vision-text latent space alignment between $z_a$ and $z_t$ and a orthogonality loss $\mathcal{L}_{orth}$ to encourage the orthogonality of $z_v$ (Sec.~\ref{sec:align and uniformity}). Visualization of the first 3 dominant dimensions of latent space $z_v$ via PCA~\cite{wold1987principal} shows the M-FLAG alleviates the dimensional collapse in the latent space, while MGCA~\cite{mgca} and GLoRIA~\cite{huang2021gloria} suffer the problem to different extents.}
\label{fig1}
\end{figure}
The proposed M-FLAG is a simple and light VLP framework that aims to learn visual and text representations by leveraging both medical images and radiology reports. We employ a \textit{freeze} strategy for the text encoder $E_T$ to mitigate ambiguity in vision-text latent space alignment. Additionally, we explicitly optimize the latent space geometry using a orthogonality loss. By doing so, we encourage the visual latent space to keep a stable geometry and reduce the risk of collapse. Fig.~\ref{fig1} illustrates the workflow of M-FLAG and the learned latent space geometry compared to two recent medical VLP approaches.\\
\noindent \textbf{Vision encoder and frozen text encoder:}
The paired medical image and text are denoted as $x_v, x_t$, respectively.  As illustrated in Fig.~\ref{fig1},  we obtain the image embedding $z_v$ via the vision encoder $E_{V}$ and the text embedding $z_t$ via the frozen text encoder $E_{T}$.\\
\noindent \textbf{Vision embedding:}
$E_{V}$, the vision embedding $z_v \in \mathbb{R}^{B \times N}$ is extracted from the last pooling layer of $E_{V}$. $N$ denotes the dimension of the latent space and $B$ represents the batch size.\\
\noindent \textbf{Text embedding:}
A text encoder $E_{T}$ extracts text embedding of word tokens from a medical report. Similar to BERT~\cite{devlin2018bert}, a special token $[cls]$ is added, which aggregates the representations of all word tokens into one embedding. The embedding of $[cls]$ is used as the text report representation and denoted as $z_t$.

\subsection{Frozen Language Model}
In this work, we use a \textit{frozen} text encoder $E_T$, which can be obtained from any general language model. The latent space of $z_v$ is thus stable without the risk of latent space perturbation~\cite{quan2022deep,zhou2019review} due to the joint training of two encoders. Naturally, the computational cost is considerably reduced since the proposed approach only requires the training of
a light vision encoder $E_V$ and a projector $p(\cdot)$. 

\subsection{Alignment and Uniformity}
\label{sec:align and uniformity}
As illustrated in Fig.~\ref{fig1}, after obtaining the visual embedding $z_{v} = E_{V}(x_v)$ and text embedding $z_t=E_{T}(x_t)$ using corresponding encoders, the vision embedding $z_v$ is projected to $z_a$ by a linear projector $z_{a}=p(z_{v})$, so that $z_a$ is of the same dimension as $z_t$ and alignment can be performed.

We compute a composite loss $\mathcal{L}_{total}$ to train the vision encoder $E_{V}$ and the projector $p(\cdot)$, which consists of two parts, alignment loss $\mathcal{L}_{align}$ and orthogonality loss $\mathcal{L}_{orth}$:
\begin{align}
    \mathcal{L}_{total} &= \mathcal{L}_{align}+\mathcal{L}_{orth} \\
    \mathcal{L}_{align} &= ||{\Bar{z_{a}}}-\Bar{{z_{t}}}||^{2}_{2} = 2-2 \Bar{{z_{a}}}^T,\Bar{{z_{t}}}\\
    \mathcal{L}_{orth} &= \sum_{i=1}\big(1- (\Bar{z_{v}}^{T}\cdot \Bar{z_{v}})_{ii}\big)^{2}   + \sum_{i \neq j}(\Bar{z}_{v}^{T}\cdot \Bar{z}_{v})_{ij}^2, \label{eq1}
\end{align}
where $\{i,j\}\in\{1,...,\textrm{dim}(z_v)\}^{2}$.
We implement $\ell_2 \text {-normalization }$ on $z_a,z_t,z_v$ to obtain $\Bar{z_a},\Bar{z_t},\Bar{z_v}$.
$\mathcal{L}_{align}$ minimizes the discrepancy between $\Bar{z_a}$ and $\Bar{z_t}$, while $\mathcal{L}_{orth}$ maximizes the independence among latent features in $\Bar{z_v}$, forcing its empirical correlation matrix to be an identity matrix. In other words, we expect different latent feature dimensions to be independent. The objective of the first term on the right side in Eq.~\eqref{eq1} aims to optimize the diagonal elements of the empirical correlation matrix to 1, while the second term on the right side aims to reduce all non-diagonal elements to 0. Here, $(\cdot)^T$ denotes the matrix transpose operation.

\section{Experiments}
\subsection{Dataset for Pre-training} 
M-FLAG is pre-trained on the MIMIC-CXR (MIMIC) dataset~\cite{johnson2019mimic,johnson2019mimicjpg}, which contains 227,827 image-text pairs with chest X-ray (CXR) images and radiology reports. Following the preprocessing procedure of~\cite{convirt,huang2021gloria,mgca}, 213,384 image-text pairs are used. We use ResNet50~\cite{resnet} as $E_V$ and frozen CXR-BERT~\cite{boecking2022making} as $E_T$. Pre-training takes 100 epochs on 8 A100 GPUs, with a batch size of 128 for each GPU and a learning rate of 0.001 using the LARS~\cite{lars} optimizer.

\subsection{Datasets for Downstream Tasks}
The pre-trained model is evaluated on 3 downstream tasks across 5 datasets:\\
\noindent \textbf{Medical image classification} is implemented on MIMIC, CheXpert (CXP), and NIH~\cite{johnson2019mimicjpg,irvin2019chexpert,wang2017chestx} datasets, each consisting of images from 14 disease categories. 
To reduce sampling bias and maintain consistency, we follow the dataset split in CheXclusion~\cite{seyyed2020chexclusion} and evaluate the macro AUC scores.\\
\noindent \textbf{Image segmentation} is evaluated on two datasets, RSNA~\cite{rsna} (pneumonia segmentation) and SIIM~\cite{siim} (pneumothorax segmentation). Following ~\cite{huang2021gloria,mgca}, we use U-Net~\cite{unet} as the segmentation backbone. The pre-trained model is used as the frozen encoder of the U-net~\cite{unet} and we only update the decoder of the U-net during fine-tuning. We evaluate segmentation performance using Dice scores.\\
\noindent \textbf{Object detection} is implemented on the RSNA~\cite{rsna} dataset for pneumonia detection, using the preprocessing techniques outlined in~\cite{mgca}. Following~\cite{mgca}, we use YOLOv3~\cite{redmon2018yolov3} as the detection framework. We employ the pre-trained vision encoder of M-FLAG as the backbone and only fine-tune the detection head. The detection task is evaluated using mean average precision (mAP) with the intersection of union (IoU) thresholds ranging from 0.4 to 0.75. 

Tab.~\ref{tab:split} reports the data split details. For all downstream tasks, we fine-tune using ${1\%, 10\%, 100\%}$ of the train data on a single A100 GPU.

\begin{table}[ht!]
\centering
\caption{Datasets are split following \cite{huang2021gloria,seyyed2020chexclusion,mgca}.}
\label{tab:split}
\scalebox{1.0}{
\begin{tabular}{lccccc}
\hline
 Task & Dataset & Split & Train & Valid & Test \\ \cline{1-6} 
\multicolumn{1}{l}{\multirow{3}{*}{\begin{tabular}[l]{@{}c@{}}Classification\end{tabular}}} & MIMIC~\cite{johnson2019mimicjpg} & ~\cite{seyyed2020chexclusion} & 215,187  & 5,000 & 23,137 \\
\multicolumn{1}{l}{} & CXP~\cite{irvin2019chexpert} & ~\cite{seyyed2020chexclusion} & 167,185 & 5,000 & 19,027 \\
\multicolumn{1}{l}{} & NIH~\cite{wang2017chestx} & ~\cite{seyyed2020chexclusion} & 100,747 & 5,000 & 6,373 \\ \hline

\multirow{2}{*}{\begin{tabular}
[l]{@{}c@{}}Segmentation\end{tabular}} & \multicolumn{1}{c}{RSNA~\cite{rsna}} &  \cite{huang2021gloria,mgca} & 16,010 & 5,337 & 5,337 \\
 & \multicolumn{1}{l}{SIIM~\cite{siim}} & \multicolumn{1}{c}{\cite{huang2021gloria,mgca}} & \multicolumn{1}{c}{8,433} & \multicolumn{1}{c}{1,807} & \multicolumn{1}{c}{1,807} \\ \hline
 
Detection & \multicolumn{1}{l}{RSNA~\cite{rsna}} &\multicolumn{1}{c}{\cite{mgca}} & \multicolumn{1}{c}{16,010} & \multicolumn{1}{c}{5,337} & \multicolumn{1}{c}{5,337} \\ \hline
\end{tabular}
}
\end{table}

\subsection{Results}
\noindent\textbf{Medical image classification:} The AUC scores on MIMIC, CXP, and CXR14 are reported in Tab.~\ref{tab: ft cls}.
It shows that M-FLAG consistently outperforms all baseline methods across almost all datasets and data fractions. Notably, M-FLAG achieves superior performance while using only 22\% of the trainable parameters compared to other methods. While MGCA~\cite{mgca} slightly outperforms our method only when fine-tuning on 10\% of the CXP dataset, it requires more than five times parameters than M-FLAG. 
\\
\noindent\textbf{Segmentation and object detection:} 
Tab.~\ref{tab:seg res} shows that M-FLAG outperforms all SOTA methods across all datasets and data fractions in segmentation and detection tasks. In the segmentation task, M-FLAG achieves the highest Dice score across all fractions of both the SIIM and RSNA datasets. Interestingly, even when fine-tuned with only 1\% of the data in RSNA, M-FLAG outperforms the ImageNet pre-trained model fine-tuned with 100\% of the data. Similarly, in the object detection task, M-FLAG achieves the highest mean average precision (mAP) across all data fractions of the RSNA dataset. When fine-tuned with only 10\% of the data, M-FLAG still outperforms the ImageNet pre-trained model with 100\% fine-tuning.

These results indicate the advantages of using a frozen language model and introducing orthogonality loss during pre-training, which may yield more informative latent representations that are better suited for downstream tasks. Overall, the improvements achieved by M-FLAG across diverse downstream tasks demonstrate its effectiveness and versatility in medical image analysis.

\begin{table}[ht!]
\centering
\caption{AUC scores (\%) of image classification tasks on MIMIC, CXP, NIH datasets with 1\%, 10\%, 100\% labeled data. }
\label{tab: ft cls}
\scalebox{1.0}{
\begin{tabular}{lcccccccccc}
\hline
     & Trainable  &\multicolumn{3}{c}{MIMIC} & \multicolumn{3}{c}{CXP} & \multicolumn{3}{c}{NIH} \\ 
    Method    & parameters (M) & 1\%         & 10\%    &100\%    & 1\%        & 10\%   &100\%    & 1\%        & 10\% &100\%      \\ \hline
Random & 38.3 & 53.6 &  66.5    &  78.2   & 62.6  &   69.0    &   76.9  & 56.4 & 67.1 & 76.9 \\
ImageNet & 38.3 &  67.8 &   70.5   & 79.3  & 63.7    &  70.7 &  77.7     &   59.7  &  68.9 &  78.1 \\ \hline

ConVIRT~\cite{convirt} & 110.3 & 67.8 &  73.4      &  80.1   & 63.2 & 71.3    &  77.7   &  60.0 &  69.0 & 76.6\\

GLoRIA~\cite{huang2021gloria} & 113.1 & 67.5 & 72.6 & 80.1 & 62.9  & 69.0  &  77.8 &    60.1  & 71.2   &    77.7    \\

MGCA~\cite{mgca}   & 113.4 & 68.4 & 74.4 & \textbf{80.2} & 63.4 & \textbf{72.1} & 78.1 & 61.1 & 67.8  &  77.3        \\ \hline

M-FLAG  & \textbf{25.6} &\textbf{69.5} & \textbf{74.8} & \textbf{80.2} & \textbf{64.4} & 71.4 & \textbf{78.1} &\textbf{62.2}      & \textbf{71.6}    &  \textbf{78.7}  \\\hline
\end{tabular}}
\end{table}

\begin{table}[ht!]
\centering
\caption{Dice (\%) of segmentation tasks on SIIM, RSNA datasets. mAP (\%) of detection task on RSNA dataset. All tasks are fine-tuned with 1\%, 10\%, 100\% labeled data. }
\label{tab:seg res}
\scalebox{1.0}{
\begin{tabular}{lcccccc|ccc}
\hline
     & \multicolumn{6}{c}{Segmentation} & \multicolumn{3}{c}{Object Detection} \\ 
 & \multicolumn{3}{c}{SIIM(Dice\%)} & \multicolumn{3}{c}{RSNA(Dice\%)} & \multicolumn{3}{c}{RSNA(mAP\%)} \\
Method & 1\% & 10\% & 100\% & 1\% & 10\% & 100\% & 1\% & 10\% & 100\% \\ \hline
Random & 9.0 & 28.6 & 54.3 & 6.9 & 10.6 & 18.5 & 1.0 & 4.0 &  8.9 \\
ImageNet & 10.2 & 35.5 & 63.5 & 34.8 & 39.9 & 64.0 & 3.6 & 8.0 &  15.7 \\ \hline
ConVIRT~\cite{convirt} & 25.0 & 43.2 & 59.9 & 55.0 & 67.4 & 67.5 & 8.2 & 15.6 & 17.9 \\
GLoRIA~\cite{huang2021gloria} & 37.4 & 57.1 & 64.0 & 60.3 & 68.7 & 68.3 &  11.6 & 16.1 & 24.8 \\
MGCA~\cite{mgca} & 49.7 & 59.3 & 64.2 & 63.0 & 68.3 & 69.8 & 12.9 &  16.8 & 24.9 \\ \hline
M-FLAG & \textbf{52.5} & \textbf{61.2} & \textbf{64.8} & \textbf{64.6} & \textbf{69.7} &\textbf{70.5} & \textbf{13.7} & \textbf{17.5} &\textbf{25.4} \\ \hline
\end{tabular}
}
\end{table}

\subsection{Dimensional Collapse Analysis}
\begin{figure}[ht!]
    \centering
    \includegraphics[width=0.99\linewidth]{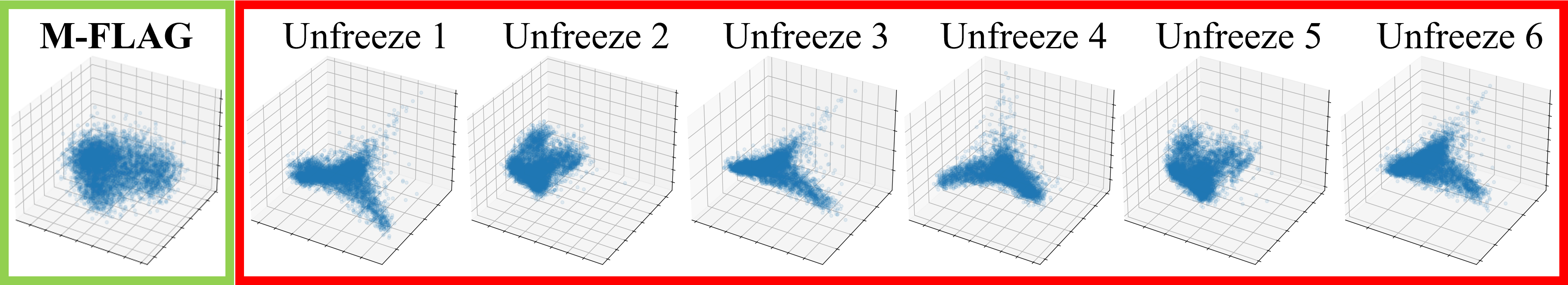}
    \caption{Visualization of the first 3 dominant PCA dimensions of latent space on NIH dataset. M-FLAG (green) is compared to its variants (red) when the last $n$ layers of the language model are not frozen.}
    \label{fig: vis unfre}
\end{figure}

Recent studies~\cite{wang2020understanding,jing2021understanding} have highlighted that latent space learned via self-supervised learning can suffer from issues such as complete collapse or dimensional collapse, which would lead to poor performance for downstream tasks. Fig.~\ref{fig1} bottom right panel shows that both MGCA and GLoRIA~\cite{mgca,huang2021gloria} suffer from dimensional collapse. Fig.~\ref{fig: vis unfre} shows that if the last $n$ layers of the language model in M-FLAG are not frozen, the latent space geometry would also exhibit varying degrees of collapse. This indicates the importance of using a frozen language model. Quantitative results in Tab.~\ref{tab: ft cls},\ref{tab:seg res},\ref{tab: ablation loss},\ref{tab:abla lm} and qualitative visualization in Fig.~\ref{fig1},\ref{fig: vis unfre} further demonstrate that a collapsed latent space can impair the performance for various downstream tasks, especially for segmentation and detection. These findings highlight the usefulness of a frozen language model in preventing latent space collapse. 

\subsection{Ablation Study}
\begin{table}[ht!]
\centering
\caption{Performance for ablation study of M-FLAG. ``only $\mathcal{L}_{orth}/\mathcal{L}_{align}$" indicates that M-FLAG is pre-trained only with $\mathcal{L}_{orth}/\mathcal{L}_{align}$.}
\label{tab: ablation loss}
\scalebox{1.0}{
\begin{tabular}{lccc|c|c}
\hline
 & MIMIC & CXP & NIH &SIIM &RSNA\\
& AUC(\%) & AUC(\%) & AUC(\%) & Dice(\%) & mAP(\%)\\
  Method & 1\% & 1\% & 1\% & 1\% & 1\% \\ \hline
only $\mathcal{L}_{align}$  & 69.3 & 62.6 & 61.4  & 45.7 & 12.1\\
only $\mathcal{L}_{orth}$  & 68.6 & 61.5 & 61.2 & 50.5 & 13.2 \\

M-FLAG  & \textbf{69.5} & \textbf{64.4} & \textbf{62.2} & \textbf{52.5} & \textbf{13.7}\\ \hline
\end{tabular}
}
\end{table}

\noindent\textbf{Ablation study:} Tab.~\ref{tab: ablation loss} presents the results of an ablation study to evaluate the impact of $\mathcal{L}_{orth}$ and $\mathcal{L}_{align}$ on model performance. Across all experiments, the proposed version of M-FLAG achieves the highest performance, with a clear advantage over implementations that only use $\mathcal{L}_{orth}$ or $\mathcal{L}_{align}$ in pre-training.
The performance of the model pre-trained with only $\mathcal{L}_{align}$ drops dramatically in segmentation and detection tasks, although less severe in the classification tasks. On the other hand, the model pre-trained with only $\mathcal{L}_{orth}$ does not suffer severe performance drop across the three tasks, indicating that the uniform latent space could have a considerable contribution to the performance of M-FLAG. Overall, these results underscore the importance of both loss functions in M-FLAG and highlight their complementary contributions.\\

\noindent\textbf{Comparing M-FLAG with frozen vs. unfrozen language models:} We conducted further experiments to evaluate the performance of M-FLAG while unfreezing the last few layers of the language model. This not only increases the number of trainable parameters but also influences the model performance. Tab.~\ref{tab:abla lm} shows that when the language model is unfrozen, the performance slightly drops, compared to M-FLAG with the frozen language model (proposed). M-FLAG achieves a better performance with an average improvement of 2.18\% than its $\text{Unfreeze}_{1-6}$ variants on the NIH dataset and an average improvement of 4.32\% on the SIIM dataset.
\begin{table}[ht!]
\centering
\caption{Performance of M-FLAG compared to its unfrozen variants. $\text{Unfreeze}_{n}$ indicates that the last $n$ layers of the language model are unfrozen.}
\label{tab:abla lm}
\scalebox{1.0}{
\begin{tabular}{lcccc|c|c}
\hline
                     & & MIMIC & CXP & NIH &SIIM &RSNA   \\ 
                     & Trainable   & AUC(\%) & AUC(\%) & AUC(\%) & Dice(\%) & mAP(\%)\\
  Method       & Parameters(M)  & 1\% & 1\% & 1\% & 1\% & 1\% \\ \hline
$\text{Unfreeze}_{1}$ & 32.6 & 67.8 & 63.1 & 59.7 & 47.2 & 12.5 \\
$\text{Unfreeze}_{2}$ & 39.7 & 68.7 &  63.3 & 60.6 & 48.9 & 12.3 \\
$\text{Unfreeze}_{3}$ & 46.8 & 68.8 & 63.7 &  60.7 & 45.8 & 10.7 \\
$\text{Unfreeze}_{4}$ & 53.9 & 68.7 & 62.6 &  60.1 & 50.3 & 11.4 \\
$\text{Unfreeze}_{5}$ & 60.9 & 68.2 & 64.1 &  59.2 & 46.8 & 11.8 \\
$\text{Unfreeze}_{6}$ & 68.1 & 68.2 & 63.7  & 59.9 & 50.1 & 11.5 \\ \hline
M-FLAG              &\textbf{25.6}&\textbf{69.5}&\textbf{64.4}&\textbf{62.2}&\textbf{52.5}&\textbf{13.7} \\ \hline
\end{tabular}
}
\end{table}

\section{Conclusion}
Simple architecture means low computational cost and stable training. In this work, we propose a simple and efficient VLP framework that includes a frozen language model and a latent space orthogonality loss function. Extensive experiments show that M-FLAG outperforms SOTA medical VLP methods with 78\% fewer parameters. M-FLAG also demonstrates its robustness by achieving the highest performance when transferred to unseen test sets and diverse downstream tasks for medical image classification, segmentation, and detection. This indicates the benefits of freezing the language model and regularizing the latent space. The results exhibit promising potential for improving the pre-training of vision-language models in the medical domain. In addition, the latent space geometry explored in this work provides useful insight for future work in VLP.

\section*{Acknowledgement}
C. Liu and R. Arcucci were supported in part by EPSRC grant EP/T003189/1 Health assessment across biological length scales for personal pollution exposure and its mitigation (INHALE), EPSRC Programme Grant PREMIERE (EP/T000414/1). W. Bai and M. Qiao were supported by EPSRC Project Grant DeepGeM (EP/W01842X/1). A. Shah was supported by a MRC Clinical Academic Research Partnership award  (MR/TOO5572/1) and by an MRC centre grant MRC (MR/R015600/1).
\clearpage
%
%
%
%
\bibliographystyle{splncs04}
\bibliography{paper3709}
\end{document}